# Automatic Incident Classification for Big Traffic Data by Adaptive Boosting SVM

Li-Li Wang[1], Henry Y.T. Ngan[1], Nelson H.C. Yung[2].

[1]Department of Mathematics, Hong Kong Baptist University, Kowloon Tong, Hong Kong

[2]Department of Electronic & Electrical Engineering, The University of Hong Kong, Hong Kong

Email: [1]llwang_hk@126.com, [1]ytngan@hkbu.edu.hk, [2]ypl.nyung@gmail.com

**Abstract**

Modern cities experience heavy traffic flows and congestions regularly across space and time. Monitoring traffic situations becomes an important challenge for the Traffic Control and Surveillance Systems (TCSS). In advanced TCSS, it is helpful to automatically detect and classify different traffic incidents such as severity of congestion, abnormal driving pattern, abrupt or illegal stop on road, etc. Although most TCSS are equipped with basic incident detection algorithms, they are however crude to be really useful as an automated tool for further classification. In literature, there is a lack of research for Automated Incident Classification (AIC). Therefore, a novel AIC method is proposed in this paper to tackle such challenges. In the proposed method, traffic signals are firstly extracted from captured videos and converted as spatial-temporal (ST) signals. Based on the characteristics of the ST signals, a set of realistic simulation data are generated to construct an extended big traffic database to cover a variety of traffic situations. Next, a Mean-Shift filter is introduced to suppress the effect of noise and extract significant features from the ST signals. The extracted features are then associated with various types of traffic data: one normal type (inliers) and multiple abnormal types (outliers). For the classification, an adaptive boosting classifier is trained to detect outliers in traffic data automatically. Further, a Support Vector Machine (SVM) based method is adopted to train the model for identifying the categories of outliers. In short, this hybrid approach is called an Adaptive Boosting Support Vector Machines (AB-SVM) method. Experimental results show that the proposed AB-SVM method achieves a satisfied result with more than 92% classification accuracy on average.

1. Introduction

In modern cities, traffic conditions are changing every moment and a single anomaly will affect the daily operations of transport and logistic corporations alike. As traffic data is collected from surveillance sensors such as digital video cameras and loop detectors on road networks, the data size is massive and easily contaminated with noise and errors. Most TCSS is controlled by human operators and an automated monitoring and response system is increasingly in need. As traffic situation changes rapidly, incident detection and classification becomes a necessity in TCSS. An ideal TCSS should be capable to carry out an AIC to differentiate various traffic incidents as well as hardware errors and transmission noise. In a simple sense, all normal traffic situations and error-free data can be considered as inliers, while abnormal traffic incidents such as congestions and vehicle stoppages as well as data errors can be regarded as outliers. Therefore, the research problem can simply focus on determining if any outlier exists in traffic data, and classifying them automatically.

In recent years, studies have been published for outlier detection (OD) [1-5]. In general, OD is to detect any datum which is deviated beyond a certain range from the majority of data. In [6, 7], the purpose of an OD is to identify data points appearing inconsistent with the majority of the data (inliers). OD is important as the outliers indicate potential abnormalities in many areas, for example, [1] suggested aircraft engine rotation defect detection, heart-rate monitors and fabric defect detection, or [3] mentioned traffic abnormality detection. There are three fundamental approaches for OD including unsupervised, semi-supervised and supervised ones. The common unsupervised methods include clustering-based method [8], distance-based method [9], and density-based method [10]. The clustering-based approach [8] defines an observation as an outlier if it deviates from the overall clustering pattern. The distance-based approach [9] assumes an observation is an outlier if the distances of a certain percentage of samples from a datum are larger than a given threshold. In the density-based approach [10], an observation is detected as an outlier if its local density is low. However, as mentioned, outliers may arise from different reasons in traffic data, such as traffic incidents, congestions due to peak hours, small volumes of vehicles, or data-capturing hardware failures. So far, researchers are interested to discover the unknown but meaningful categories of outliers, although most of these methods are not able to distinguish the types of anomalies. In fact, each type of outliers has its own characteristics. If we could identify the category of the current traffic state, this piece of information could be useful for traffic control. To identify the category of each outlier, one can resort to supervised learning approach [11-15]. This approach requires a pre-labeling of data as normal or abnormal which could be further classified into different categories. Based on these labeled training data, a supervised learning method is then adopted to a certain classifier to identify the category of an outlier after learning. In the literature, SVM [14] and adaptive boosting techniques [15] are the most popular methods and have achieved better classification performance than others. SVM as a supervised learning model which is widely employed for the aim of classification in machine learning. In [16], SVM-based method was successfully applied to recognize daily activity pattern for the forecasting purpose of travel demand. Adaptive boosting technique [15] as an ensemble method is capable of training a strong classifier by combining a series of moderately accurate component classifiers. This method has been proven [15] to boost classification accuracy.

In order to identify certain categories of traffic incidents based on these learning methods, a massive traffic database for creating one or more classifiers is required. In order to achieve that, traffic data is collected via multiple sensors and is considered as big data by nature. In this research, traffic data were first collected from surveillance video cameras over 31 days. These traffic data were then converted into ST signals. These ST signals generally present high ST similarity within signals or among signals in the same period of each day. For evaluation purpose, an extension of the database with a simulated process of the traffic data is performed. Details of the above will be given in Section 2.2.

Note that for abnormalities, they show different characteristics, and their quantity is generally assumed to be less than that of the normal data. The collected data sets are thereby imbalanced in terms of the number of samples available. If the imbalanced data set is used for learning, the performance of the learning algorithm(s) would degrade significantly [17], as most methods tend to build the classifiers from the majority category of data. As a result, the predictive accuracy is usually higher. However, the identification rate of the minority class is quite low. In extreme cases, all testing samples may be mis-classified as the majority category. This would be meaningless in an OD application. Researchers have come up with remedies for this imbalanced data problem. The commonly used methods include performing data re-distribution [18] or classifier modification [19]. Performing data re-distribution means resampling the majority class or generating simulated minority class to achieve a more balance weighs for different data categories. Modifying the design of a classifier to adapt data characteristics is another choice to deal with the imbalanced data problem.

Therefore, in this paper, we target to learn the daily traffic patterns at a four-arm junction by combining both AdaBoost and SVM (AB-SVM) methods. The goal of the proposed AB-SVM method is to not only identify any outliers that show inconsistency with the majority of traffic data, but also distinguish their categories for investigating detailed and useful information from it. The keys to AIC lie in determining which traffic flow signal representation is useful and meaningful in order to identify an incident type as well as in dealing with imbalance data as a whole. In brief, an AdaBoost method is firstly used to classify the imbalanced data as inliers or outliers. Afterward, a SVM model is trained up to classify the categories of outliers by using only abnormal ST signals. The advantages of this proposed strategy are that (i) it can reduce the training complexity compared with multi-class AdaBoost, (ii) data imbalance problem can be effectively abated by the hybrid AB-SVM techniques, and (iii) the categories of outliers are identified.

The rest of the paper is organized as follows. Related work including the generation method of big simulated traffic data, feature representation and review of AIC methods are given in Section 2. The proposed AB-SVM method to detect outliers and identify their categories is presented in Section 3. Experimental results are given and discussed in Section 4. Section 5 concludes the paper.

## 2. Related work

This section discusses the collection of ST signals, the generation method of big simulated traffic data, signal representation method, and the existing methods of AIC on traffic database.

### 2.1 Extraction of ST signals

To identify traffic behaviors for the need of TCSS, massive traffic data collections for feeding machine learning classifiers are required. Fig. 1 shows the flowchart of traffic data collections. Details are introduced in this section.

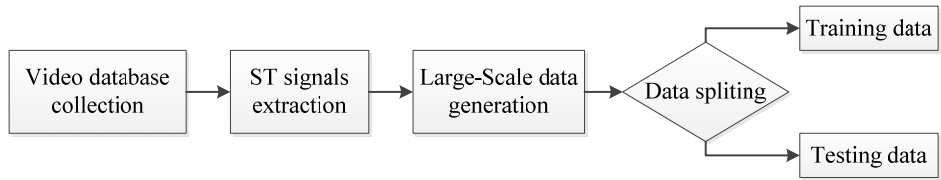

Fig.1 Generation of big traffic database

The traffic video database employed in this paper was collected from a four-armed junction in Hong Kong as shown in Fig.2. The video database was recorded for 31 days from December 28, 2010 to January 27, 2011 with two sessions per day: 07:00-10:00 AM and 17:00-20:00 PM. According to the motion state of the traffic flow controlled by an array of traffic lights, the traffic flows of this 4-arm junction is characterized by four motion patterns (MPs) as shown in Fig.3. During the period of each MP, a traffic flow volume in its respective traffic direction is recorded. For example, when the current MP as shown in Fig. 2 (a) happens, four traffic flow volumes $\{w_i, i = 1, 2, 3, 4\}$ are recorded in four corresponding directions.

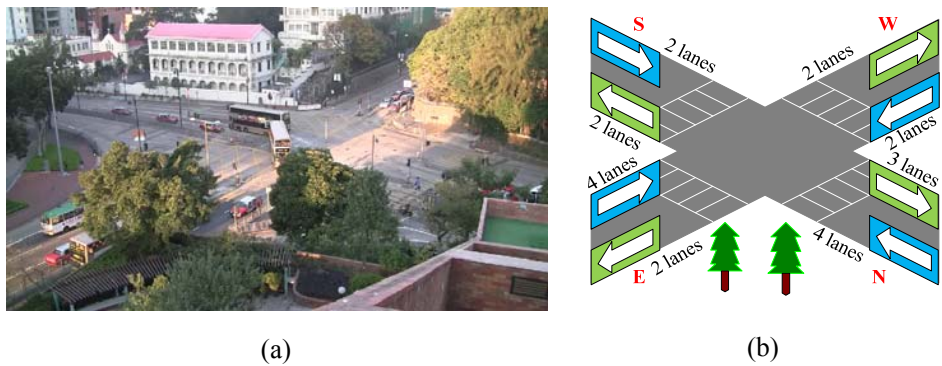

(a)          (b)

Fig.2 (a) Real scene of the 4-arm junction and (b) ideal map of the junction

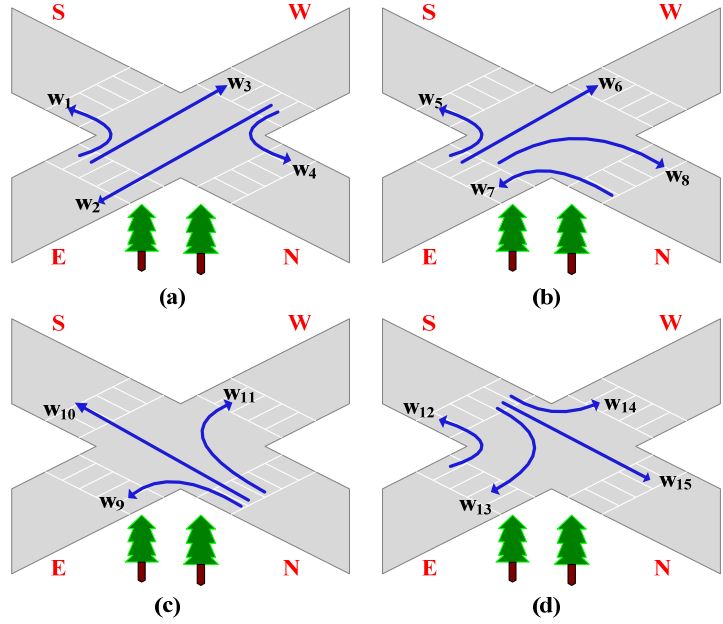

Fig.3 Four MPs for vehicle counting: (a) MP1, (b) MP2, (c) MP3 and (4) MP4

In our research, we extract 19 ST signals from 19 traffic directions for each session to characterize the traffic conditions of this 4-arm junction for AIC. More specifically, there are 4 entry, 4 exit and 11 entry direction distribution (EDD) signals. To have consistent representation, we truncate 80 traffic cycles for each session with 3 hours. In other words, each ST signal is represented by this 3-hour session length (i.e. 80 cycles) as shown in Fig. 4(a). Due to an inconsistency of traffic flow behavior in weekends of the 31-day data, only data on weekdays (23 days in total) are used in our research. For the 4-arm junction, traffic signals are expressed as $\{z_i, i=1,…, 19\}$ for 19 traffic directions. The signal in each cycle is a summation of traffic flow volumes (i.e. number of vehicle passing through) along its direction from all four traffic MPs. The 23 signals collected from each direction in 31 days are then individually analyzed to learn about the traffic behavior. Mathematically, the relationships between the signals and the MPs $\{m_j | j = 1, ... , 4\}$ are denoted as

Table 1. Mathematical relationships between signals and MPs

$$z_1 = f(m_j | j = 1,2,4) = \sum_{i=1,3,5,6,8,12} w_i$$

$$z_2 = f(m_4) = \sum_{i=13,14,15} w_i$$

$$z_3 = f(m_1) = \sum_{i=2,4} w_i$$

$$z_4 = f(m_j | j = 2,3) = \sum_{i=7,9,10,11} w_i$$

$$z_5 = f(m_j | j = 1, ...,4) = \sum_{i=2,7,9,13} w_i$$

$$z_6 = f(m_j | j = 1, ...,4) = \sum_{i=1,5,10,12} w_i$$

$$z_7 = f(m_j | j = 1, ...,4) = \sum_{i=3,6,11,14} w_i$$

$$z_8 = f(m_j | j = 1,2,4) = \sum_{i=4,8,15} w_i$$

$$z_9 = f(m_j | j = 1,2,4) = \sum_{i=1,5,12} w_i$$

$$z_{10} = f(m_2) = w_8$$

$$z_{11} = f(m_j | j = 1,2) = \sum_{i=3,6} w_i$$

$$z_{12} = f(m_4) = w_{14}$$

$$z_{13} = f(m_4) = w_{13}$$

$$z_{14} = f(m_4) = w_{15}$$

$$z_{15} = f(m_1) = w_4$$

$$z_{16} = f(m_1) = w_2$$

$$z_{17} = f(m_i | i = 2,3) = \sum_{i=7,9} w_i$$

$$z_{18} = f(m_3) = w_{11}$$

$$z_{19} = f(m_3) = w_{10}$$

According to our investigation on captured videos, the extracted traffic data may be grouped into three obvious categories: normal traffic data (inlier), abrupt low traffic volume state due to congestion or hardware failure (outlier), repeated jams (outlier). Fig. 4 depicts some examples of the three categories. Each category has its own characteristic as follows:

Table 2 Categories of traffic signals

| Category | Inlier/Outlier | Definition | Characteristic of ST signals | Figure |
|---|---|---|---|---|
| 1 | Inlier | a) smooth traffic flow in the investigated direction b) vehicles pass through this intersection without any delay | Steady wave pattern | Fig. 4(a) |
| 2 | Outlier: slight jam | a) traffic flow is broken abruptly in the investigated direction b) vehicles pass through the intersection slowly due to slight traffic jams | Abrupt low traffic volumes over 3 cycles | Fig. 4(b) |
| | Outlier: hardware failure | surveillance equipment found errors | Hardware error with low traffic volumes near to zero | Fig. 4(c) |
| 3 | Outlier: serious jam | a) low traffic flow in the investigated direction b) vehicles pass through the intersection slowly due to serious traffic jams | Repeated low traffic volumes, may have an incident also | Fig. 4(d) |

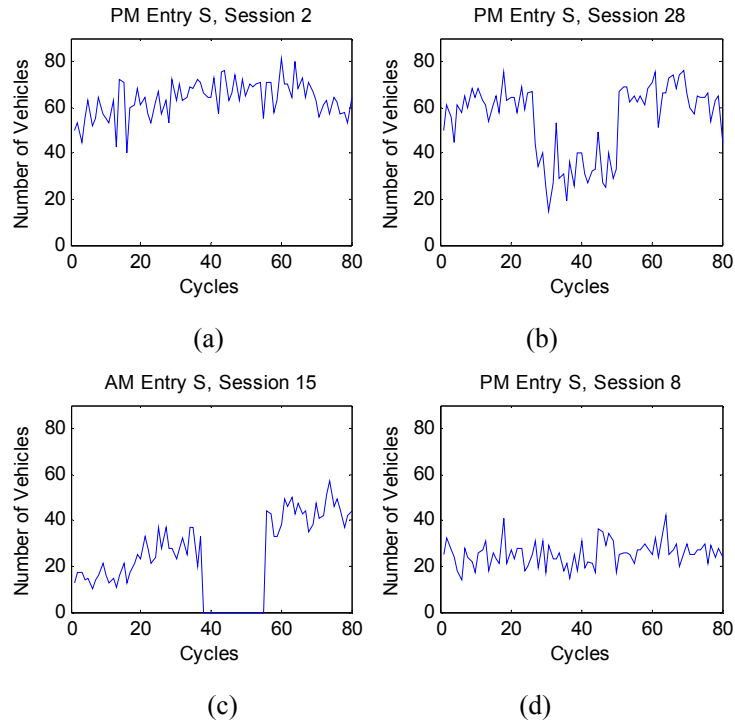

Fig.4 Categories of traffic data: (a) normal traffic ST signal, (b) ST signal with jams leading to low volumes, (c) ST signal with hardware failure, and (d) ST signal with repeated jams

2.2 Generation of big simulated traffic data

As only 23 ST signals are available for each traffic direction in the original captured data, it is not enough for learning based methods. Besides, this quantity of data does not represent every possible variation of each abnormal category. As a result, it limits the generation ability of most learning based methods. Still, it is very time-consuming to collect further massive amount of traffic data to rectify the problem. To strike a balance, we build an additive model of embedding Gaussian noise under a constraint of a high signal-noise-ratio (SNR) to generate a large amount of simulated data for the purpose of training and testing. Fig. 5 shows an example of the simulated ST signals and their distribution in 2D PCA space with different values of SNR. From the figure, we can see that the fluctuation of ST signals is quite large with a small value of SNR. In 2D PCA space, they are dispersed randomly due to large noises introduced. By contrast, the generated ST signals have small fluctuation and regular distribution in 2D PCA space when a higher value of SNR is used.

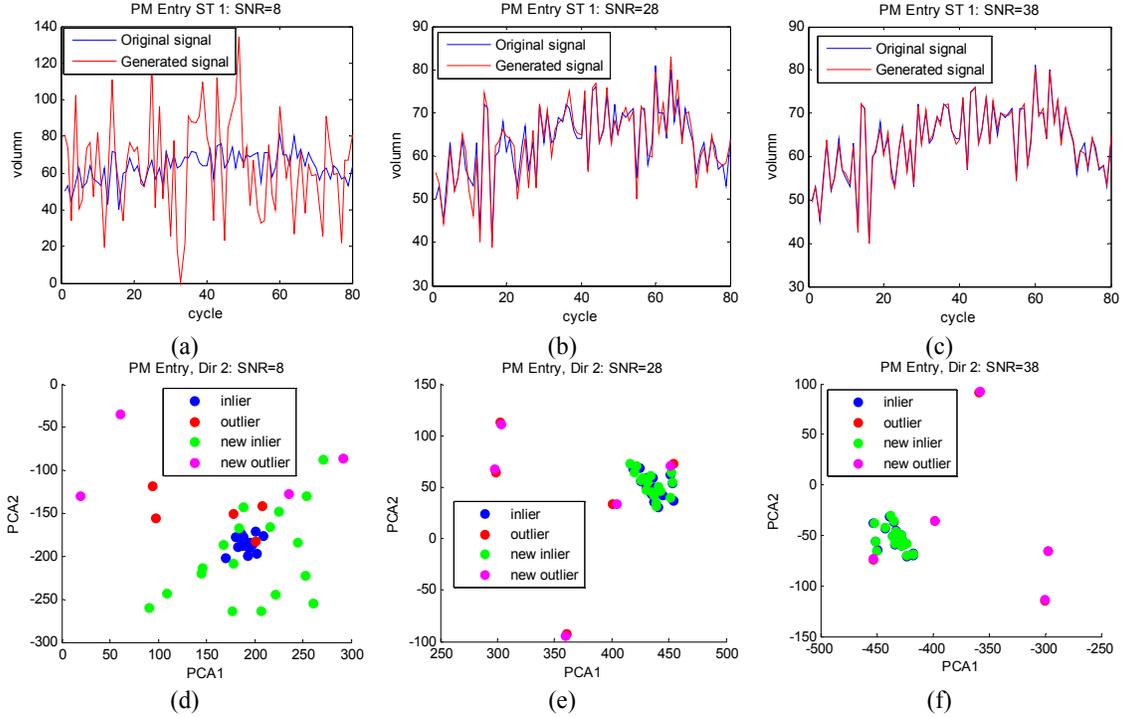

Fig.5 Generated ST signals and their distribution in 2D PCA space with different values of SNR based on the additive model

Outliers are then synthesized over several consecutive cycles randomly by resembling their characteristics as shown in Table 2. Totally, a quantity of 322 (23*14) ST signals are generated from the original 23 ST signals by adding Gaussian white noise for each direction (totally 19 directions). Take traffic direction 2 in the PM session as an example. In this direction, there are 5 abnormal and 18 normal ST traffic signals. Each ST traffic signal $s(n)$ will have a multiple of 14 to generate an overall 322 ST signals through the following procedures:

Step 1: check all 23 ST signals $\{s_i, i = 1, \ldots, 23\}$, and substitute abnormal signals with normal signals in adjacent days to form a new group of signals $\{s'_i, i = 1, \ldots, 23\}$.

Step 2: add Gaussian white noise $N_i$ to each signal $s'_i$ according to the predefined value of SNR.

$$x'_i = s'_i + N_i, \quad i = 1, \ldots, 23 \tag{1}$$

Step 3: choose M signals from $\{x'_i, i = 1, ...,23\}$, and transform them into abnormal signals (outliers) according to eqns. (2)-(3) for two categories of outliers.

$$x_i^I(n) = \begin{cases} th, & N_0 \leq n \leq N_0 + L_1 \\ x'_i(n), & otherwise \end{cases} \quad n = 1, ...,80 \quad (2)$$

$$x_i^{II}(n) = max(0, \ x'_i(n) - \alpha) \quad n = 1, ...,80 \quad (3)$$

where $L_1 \geq 4$. We assume that it is an abnormal signal belonging to Category 2 if consecutive 4 or more traffic cycles with smaller number of traffic volumes happen in one session. $N_0$ ranges from 1 to 23-L+1. According to our observation, the number of outliers in AM sessions is usually smaller than that in PM sessions. Therefore, we set M to 2 and 4 for the AM and PM sessions in our experiment, respectively.

2.3 Feature extraction of ST signals

A proper feature extraction could improve classification performance. As discussed in our previous work [20], big traffic data are easily contaminated with noise during data collection. Since these signals are very similar in the original ST domain, it is not easy to identify outliers. On the other hand, the complexity is too high if a whole piece of ST signal is directly input as one feature vector. According to our investigation, outliers in traffic data due to low vehicle volumes generally exist at least over several traffic cycles. In this paper, the mean of vehicles from four consecutive cycles is calculated. The process of mean from consecutive cycles for each ST signal is shown in Fig. 6, where $f(n) = \sum_{m=n}^{n+3} x(m)/4$, and $x(m)$ denotes the number of vehicles in the $m^{th}$ cycle. This averaging process is applied to all ST signals. The Mean-Shift filtering has the effect of eliminating signal values which are quite different from their surroundings. After accomplishing this process, the feature dimension of each signal is still high (i.e. 80 in this research). Principle Component Analysis (PCA) is well known to keep a signal quality by just extracting its main components, for which it reduces its dimensionality for feature representation as well. Herein, the 80-dimension feature vectors of each ST signal is analyzed by the PCA to keep just the first several components in order to reduce the complexity in the training and testing stages. The coefficients of the first several components, being a new domain with low-dimension, are used for AIC.

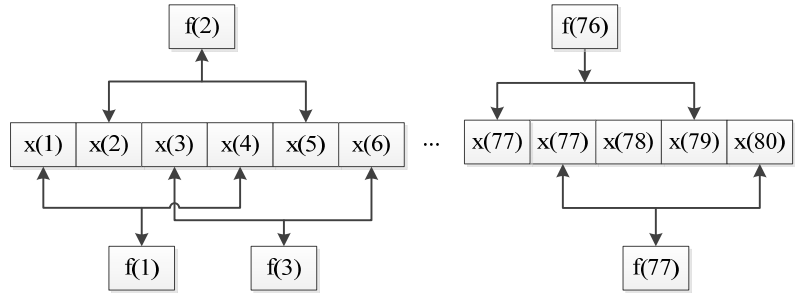

Fig. 6. Feature extraction example of one ST signal in terms of every four consecutive cycles

2.4 Popular methods to be used for AIC on big traffic database

Many countries in the world are suffering from traffic congestions in their highway and urban roads. Traffic incident is one of the major reasons [21] to traffic congestions. Delayed response to incidents would deteriorate the traffic further, and make incidents clearance and post-processing traffic more difficult. Therefore, it is indeed necessary to have an automatic and reliable incident detection and classification system. OD technique has been applied in traffic data analysis for several decades. Outlier means abnormal actions exist in traffic. Many OD algorithms have been developed for automatic incident detection (AID). Among these algorithms, machine learning algorithms [22-27] are most popular and widely investigated.

Since 1990s, machine learning techniques are widely applied for AID. Various variations of Artificial Neural Network (ANN) were firstly investigated. In [22], multi-layer feed-forward ANN (MLFANN) was used to detect incidents and showed better AID performance. In [23], Jin et. al adopted probabilistic ANN (PANN), and better detection success ratio (DSR) and false acceptance ratio FAR performance were achieved. To further improve AID performance, SVM [24-26] was investigated to detect outliers. Experimental results in [24] show that SVM can generate better AID performance than ANN. In [25], Yuan and Cheu discovered that SMVs with polynomial kernel and RBF kernel achieve high classification accuracy than MLFANN and PANN. With the development of machine learning, boosting technique was proposed to combine a set of weak learners to construct a single strong learner for classification. This concept is first proposed by Kearns and Valiant in 1988 and 1989 [27, 28]. At present, there are many boosting algorithms. The initial one was proposed by Freund in 1995 [29]. However, this method cannot support the subsequent weak learners adaptive to the instances which are misclassified by previous classifiers. In order to adapt to the weak learners, adaptive boosting (AdaBoost) algorithms, such as Real AdaBoost [30], Gentle AdaBoost [31] and others, were developed. Due to the outstanding performance of adaptive boosting methods, they have been applied to solve imbalanced dataset problems [32].

Note that the above researches only focus on outlier detection in traffic data, while detailed incident categories are also very useful for traffic management and optimization of traffic road in a city. In this paper, we take advantages of AdaBoost method to detect outliers in imbalanced traffic dataset, and further make full use of SVM to learn the abnormal traffic behaviors to identify their categories.

## 3. Automatic incident classification of traffic data

In general, the quantity of inliers (normal data) is usually much larger than that of the outliers (abnormal cases). However, the minority class is more interesting for the application of AIC. To discern the small number of outliers from a big database, we propose a hybrid method by taking advantages of both AdaBoost and SVM (AB-SVM) techniques to solve the classification problem of imbalanced dataset in this section. Fig. 7 depicts the flowchart of the AB-SVM classifier. The AB-SVM is developed for OD of traffic data by training a strong classifier, and further identification of the outliers categories based on trained support vectors.

More specifically, the training and testing data sets are first available based on the steps as depicted in Fig. 1. Features are then extracted based on the method in Section 2.3. PCA is subsequently adopted to reduce the feature vector dimension of the training data preceding the training process. In the AB-SVM framework, adaptive boosting (AdaBoost) technique is adopted for training a strong classifier to differentiate normal traffic signals (inlier) from abnormal ones (outliers), and SVM is employed to learn a base model from the low dimensional features of the outliers to classify their different types, such as jams leading to abrupt low volumes, repeated jams or hardware failure. If one traffic datum is detected as an outlier based on the AdaBoost classifier, the trained SVM model is further utilized to recognize the specific category of the predictive outlier. Details about this algorithm is illustrated as follows.

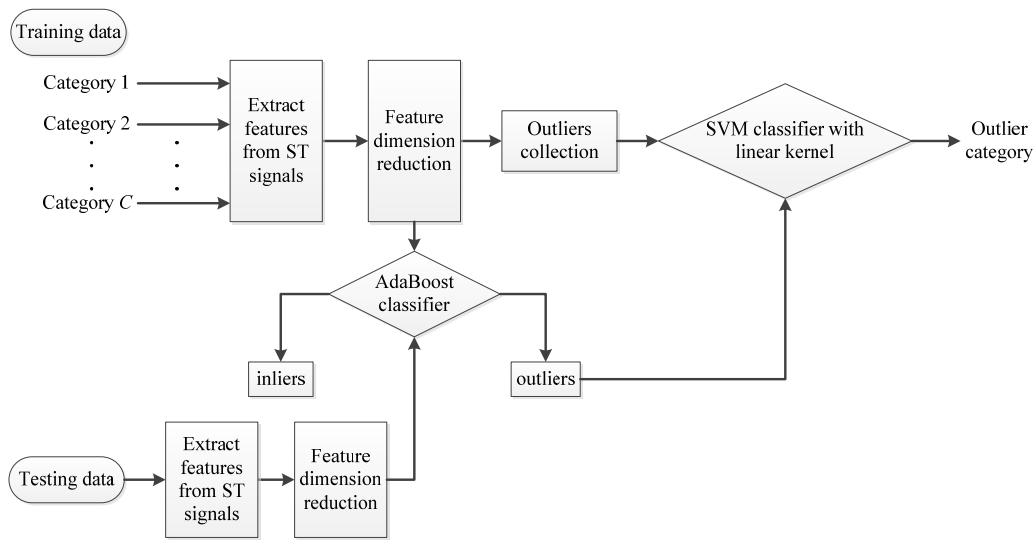

Fig.7 Flowchart of the proposed hybrid AB-SVM model

*3.1 OD based on AdaBoost*

Adaptive boosting (AdaBoost) is a particular machine learning method used to train a series of weak classifiers. During the training process for AdaBoost, the weights of the training samples are adaptively updated after each boosting iteration. The weights of the training samples which are misclassified by the current component classifier are increased, while the weights of the training samples with correct classification are decreased. Finally, the weak classifiers are combined linearly to form a strong classifier, which is expressed as

$$F(x) = \max_c \sum_{t=1}^{T} h_t(v) = \max_c \sum_{t=1}^{T} h(v, f_t, \theta_t, c) \quad (4)$$

where $h_t$ is the $t^{th}$ weak learner, $\theta$ is a threshold, $v$ denotes a feature vector in the PCA space, and $f_t$ means that the $f^{th}$ component of $v$ is used as input feature in weak learner $h_t(v)$. The training procedures of the AdaBoost classifier for OD are summarized as follows:

Step 1: Given a training set including $N_1$ positive and $N_2$ negative signals, and $N_1 + N_2 = N$.

$$(x_1, y_1), \ldots, (x_N, y_N)$$

where $\{x_i\}_{i=1}^N$ is the set of input signals in $R^D$, and $y_i$ denotes the label of input signals.

Step 2: Repeat for $c = 0, 1$

Step 2.1 Initialize the weights $w_{1,i}^c$ ($w_{1,i}^c = \frac{1}{2N_1}, \frac{1}{2N_2}$ for $d_i = 0, 1$) of training samples. The weight of each training sample is inversely proportion to the number of samples in its own group. $\{w_{1,i}^c\}_{i=1}^N$ denotes a probability distribution of training examples.

Step 2.2 For each cycle t=1,2,…T

Step 2.2.1 Normalize the weights $w_{t,i}^c$ as

$$w_{t,i}^c = \frac{w_{t,i}^c}{\sum_{j=1}^N w_{t,j}^c}, \text{ for i=1,…,N} \quad (5)$$

Step 2.2.2 Define a weak learner as $h_t(v) = h(v, f_t, \theta_t, c)$

$$h(v, f_t, \theta_t, c) = a\delta(v^f > \theta_t) + b \quad (6)$$

Step 2.2.3 Evaluate error

$$J_t(\theta) = \sum_{i=1}^N w_{t,i}^c (y_i^c - h(v_i, f_t, \theta_t, c))^2 \quad (7)$$

where $y_i^c \in \{-1, 1\}$ denotes the label of the training signal i.

Step 2.2.4 Find the best weak learner $h(v, f_t, \theta_t, c)$ with parameters $a, b, f_t, \theta_t$ through minimizing $J_t(\theta)$.

Step 2.2.5 Update the classifier $H(v,c)$ for class c and weights of training samples

$$H(v,c) = H(v,c) + h(v, f_t, \theta_t, c) \tag{8}$$

$$w_{t+1,i}^c = w_{t,i}^c \exp(-h(v, f_t, \theta_t, c)) \tag{9}$$

Step 2.2.6 If t<T, increase t by 1, and go to Step 2.2.1; otherwise go to Step 2 until all classes have been checked.

Step 3: Output the final strong classifier as denoted in Eq. (4).

*3.2 Abnormal incidents classification based on SVMs*

In the field of machine learning, SVM technique [33] is widely used for classification problem. Compared with neural networks, SVM techniques are easy to be implemented and to offer satisfactory classification results in a wide variety of application domains, such as semantic image classification [34], handwritten recognition [35], and so on. In the AB-SVM, we apply SVMs [15] to classify abnormal traffic behaviors in traffic data. Experimental results show that the hybrid model achieves superior performance for AIC. Let *C* denote the set of abnormal traffic categories, and $C = \{1, ..., L\}$ where *L* denotes the total number of outlier classes. Given M training samples from abnormal traffic signals,

$$(x_1, d_1), ..., (x_M, d_M)$$

where $\{x_i\}_{i=1}^M$ is the feature set of input signals in $R^D$, and $d_i$ denotes the label of input signals. To separate different abnormal traffic categories, the maximum margin to the hyper-plane can be obtained by solving the following optimization problem during the training process.

$$\begin{cases} \min_{w,b,p} \frac{1}{2} \sum_{m \in K} \|w_m\|^2 + \gamma \sum_{i=1}^{N} \sum_{m \neq d_i}^{\|C\|} p_i^m \\ s.t.: w_{d_i}^T \cdot x_i - b_{d_i} - (w_m^T \cdot x_i - b_m) \geq 1 - p_i^m, \\ p_i^m \geq 0, i = 1, ..., n; m \in \{1, ... L\} \setminus d_i. \end{cases} \tag{10}$$

where $w_m \in R^D$ is a vector composed of weighting coefficients for class $m$, $\gamma$ is a regularization parameter, which controls the model complexity and the training error. $\gamma \sum_{i=1}^{N} \sum_{m \neq d_i}^{\|L\|} p_i^m$ is a penalty term, which is used to penalize misclassified samples, and $p_i$ denotes the distance of the sample from the margin if it is classified wrongly.

By introducing Lagrange multipliers $\{\alpha_i\}_{i=1}^{N}$ and dual transformation [36-38], the model parameters $(w, b, p, \alpha)$ are obtained. The category of abnormal traffic ST signals in the test set are predicted as

$$d_i = sign(w^T x_i) = sign(\sum_{n \in SV} \alpha_n d_n x_n x_i) = sign(\sum_{n \in SV} \alpha_n d_n k(x_n, x_i)) \quad (11)$$

where $SV$ denotes the set of support vectors, $k(x_n, x_i)$ is a kernel function, and it is the inner product of two feature vectors. By using this kernel function, the training samples can be mapped from an input space to another feature space which makes samples more separated. There are three commonly used kernels for SVMs namely linear, polynomial and radial basis function (RBF). For the application of AIC, the SVM with linear kernels achieves outstanding performance. This will be discussed in Section 4.

*3.3 Testing based on the trained AB-SVM classifier*

As a consequence, the testing stage can be carried out as depicted in Fig. 7 based on the hybrid method. Given a testing sample, the feature is first extracted and feature dimension is reduced based on PCA. The extracted feature vector with a low dimension is passed through the trained AdaBoost classifier. If it is classified as outliers, the learned SVM model is further adopted to identify its abnormal behavior. Otherwise this testing sample is a normal signal.

4. Experimental result

In this experiment, we perform the AIC for 19 directions in the four-arm junction. Each direction includes three traffic states, and 345 traffic ST signals are used in the experimental work. A 5-fold evaluation is performed on these 345 traffic ST signals including 276 for training and 69 for testing. The experiments were performed on a platform using Intel(R) Core(TM)2 Duo CUP E8500 @3.16 GHz and RAM of 4.0 GB. The proposed algorithm is coded in MATLab. It takes about 63.30 seconds and 70.02 seconds by using the proposed AIC method for the AM and PM sessions, respectively.

4.1 Classification accuracy

Tables 3 and 4 list the classification accuracies by using the proposed method with linear kernel for the AM and PM sessions, respectively. Since the features of the ST signal through PCA are more discriminant, therefore the proposed method shows good performance by using linear kernels. From the results, we can see that the proposed method achieves high classification performance (more than 92%). In regard of the average classification accuracy, the AM and PM sessions obtained 98.37% and 98.44%, respectively.

Table 3. Classification results based on the proposed method in the AM sessions (%)

|  | Direction | 1 | 2 | 3 | 4 | 5 | Average |
|---|---|---|---|---|---|---|---|
| Entry | E | 98.55 | 100 | 98.55 | 100 | 100 | 99.42 |
|  | S | 94.20 | 98.55 | 98.55 | 94.20 | 98.55 | 96.81 |
|  | W | 100 | 98.55 | 100 | 100 | 100 | 99.71 |
|  | N | 97.10 | 97.10 | 98.55 | 98.55 | 97.10 | 97.68 |
| Exit | E | 100 | 100 | 100 | 100 | 100 | 100 |
|  | S | 98.55 | 100 | 95.65 | 97.10 | 100 | 98.26 |
|  | W | 95.65 | 100 | 97.10 | 100 | 97.10 | 97.97 |
|  | N | 98.55 | 98.55 | 98.55 | 100 | 97.10 | 98.55 |
| EDD | $E_l$ | 97.10 | 100 | 100 | 98.55 | 97.10 | 98.55 |
|  | $E_r$ | 100 | 97.10 | 100 | 97.10 | 98.55 | 98.55 |
|  | $E_s$ | 98.55 | 100 | 98.55 | 100 | 100 | 99.42 |
|  | $S_l$ | 94.20 | 98.55 | 97.10 | 97.10 | 95.65 | 96.52 |
|  | $S_r$ | 100 | 98.55 | 98.55 | 100 | 98.55 | 99.13 |
|  | $S_s$ | 100 | 100 | 98.55 | 97.10 | 100 | 99.13 |
|  | $W_l$ | 95.65 | 100 | 100 | 98.55 | 98.55 | 98.55 |
|  | $W_s$ | 97.10 | 98.55 | 92.75 | 95.65 | 95.65 | 95.94 |
|  | $N_l$ | 95.65 | 98.55 | 97.10 | 100 | 98.55 | 97.97 |
|  | $N_r$ | 98.55 | 98.55 | 100 | 98.55 | 97.10 | 98.55 |
|  | $N_s$ | 100 | 97.10 | 94.20 | 100 | 100 | 98.26 |
|  | Average |  |  |  |  |  | 98.37 |

Table 4. Classification results based on the proposed method in the PM sessions (%)

|  | Direction | 1 | 2 | 3 | 4 | 5 | Average |
|---|---|---|---|---|---|---|---|
| Entry | E | 100 | 100 | 100 | 100 | 98.55 | 99.71 |
|  | S | 95.65 | 98.55 | 97.10 | 94.20 | 98.55 | 96.81 |
|  | W | 95.65 | 98.55 | 98.55 | 98.55 | 95.65 | 97.39 |
|  | N | 98.55 | 100 | 97.10 | 98.55 | 98.55 | 98.55 |
| Exit | E | 95.65 | 100 | 97.10 | 98.55 | 98.55 | 97.97 |
|  | S | 95.65 | 97.10 | 97.10 | 98.55 | 98.55 | 97.39 |
|  | W | 95.65 | 94.20 | 98.55 | 100 | 98.55 | 97.39 |
|  | N | 100 | 98.55 | 100 | 98.55 | 97.10 | 98.84 |
| EDD | $E_l$ | 100 | 100 | 100 | 100 | 98.55 | 99.71 |
|  | $E_r$ | 97.10 | 94.20 | 100 | 95.65 | 97.10 | 96.81 |
|  | $E_s$ | 100 | 100 | 98.55 | 100 | 98.55 | 99.42 |
|  | $S_l$ | 97.10 | 95.65 | 100 | 98.55 | 100 | 98.26 |
|  | $S_r$ | 100 | 100 | 100 | 100 | 100 | 100 |
|  | $S_s$ | 97.10 | 98.55 | 100 | 100 | 98.55 | 98.84 |
|  | $W_l$ | 100 | 100 | 100 | 97.10 | 100 | 99.42 |
|  | $W_s$ | 100 | 100 | 100 | 100 | 95.65 | 99.13 |
|  | $N_l$ | 97.10 | 98.55 | 98.55 | 100 | 98.55 | 98.55 |
|  | $N_r$ | 100 | 97.10 | 100 | 98.55 | 94.20 | 97.97 |
|  | $N_s$ | 98.55 | 98.55 | 95.65 | 98.55 | 100 | 98.26 |
|  | Average |  |  |  |  |  | 98.44 |

Tables 5 and 6 show the confusion matrix for all 19 traffic directions in the AM and PM sessions, respectively. From the results, we can see that inliers (normal data) are detected with a high degree of precision (almost 100%). The repeated traffic jam (category 3) is also easily recognized with an average of 99% success rate. The 1% is misclassified as slight traffic jam (category 2) in the AM session, and 1% of category 3 is misclassified as inliers in the PM session. Compared with Categories 1 and 3, slight traffic jam is more difficult to be recognized with 64% and 83% average success rates for the AM and PM sessions, respectively. For the AM session, 35% and 1% of slight traffic jam situations are misclassified as normal state (Category 1) and repeated traffic jam (Category 3), respectively. For the PM session, 15% and 2% of slight traffic jam cases are misclassified as normal state (Category 1) and repeated traffic jam (Category 3), respectively.

Table 5. Confusion matrix in the AM sessions (%)

| | | E | | | | S | | | | W | | | | N | | |
|---|---|---|---|---|---|---|---|---|---|---|---|---|---|---|---|---|
| | C | 1 | 2 | 3 | C | 1 | 2 | 3 | C | 1 | 2 | 3 | C | 1 | 2 | 3 |
| Entry | 1 | 100 | | | 1 | 100 | | | 1 | 100 | | | 1 | 100 | | |
| | 2 | 14 | 86 | | 2 | 55 | 45 | | 2 | | 100 | | 2 | 57 | 43 | |
| | 3 | | | 100 | 3 | | | 100 | 3 | | 6 | 94 | 3 | | | 100 |
| | | E | | | | S | | | | W | | | | N | | |
| | C | 1 | 2 | 3 | C | 1 | 2 | 3 | C | 1 | 2 | 3 | C | 1 | 2 | 3 |
| Exit | 1 | 100 | | | 1 | 100 | | | 1 | 100 | | | 1 | 100 | | |
| | 2 | | 100 | | 2 | 43 | 57 | | 2 | 50 | 50 | | 2 | 45 | 55 | |
| | 3 | | | 100 | 3 | | | 100 | 3 | | 13 | 87 | 3 | | | 100 |
| | | $E_l$ | | | | $E_r$ | | | | $E_s$ | | | | $S_l$ | | |
| | C | 1 | 2 | 3 | C | 1 | 2 | 3 | C | 1 | 2 | 3 | C | 1 | 2 | 3 |
| | 1 | 100 | | | 1 | 100 | | | 1 | 100 | | | 1 | 100 | | |
| | 2 | 42 | 58 | | 2 | 33 | 67 | | 2 | 13 | 87 | | 2 | 80 | 20 | |
| | 3 | | | 100 | 3 | | | 100 | 3 | | | 100 | 3 | | | 100 |
| | | $S_r$ | | | | $S_s$ | | | | $W_l$ | | | | $W_s$ | | |
| | C | 1 | 2 | 3 | C | 1 | 2 | 3 | C | 1 | 2 | 3 | C | 1 | 2 | 3 |
| EDD | 1 | 100 | | | 1 | 100 | | | 1 | 100 | | | | 100 | | |
| | 2 | 19 | 81 | | 2 | 20 | 80 | | 2 | 42 | 58 | | | 45 | 35 | 20 |
| | 3 | | 6 | 97 | 3 | | | 100 | 3 | | | 100 | | | 7 | 93 |
| | | $N_l$ | | | | $N_r$ | | | | $N_s$ | | | | *Average* | | |
| | C | 1 | 2 | 3 | C | 1 | 2 | 3 | C | 1 | 2 | 3 | C | 1 | 2 | 3 |
| | 1 | 100 | | | 1 | 100 | | | 1 | 100 | | | 1 | 100 | | |
| | 2 | 33 | 67 | | 2 | 36 | 64 | | 2 | 38 | 62 | | 2 | 35 | 64 | 1 |
| | 3 | | | 100 | 3 | | | 100 | 3 | | | 100 | 3 | | 1 | 99 |

Table 6. Confusion matrix in the PM sessions (%)

| | | E | | | | S | | | | W | | | | N | | |
|---|---|---|---|---|---|---|---|---|---|---|---|---|---|---|---|---|
| | C | 1 | 2 | 3 | C | 1 | 2 | 3 | C | 1 | 2 | 3 | C | 1 | 2 | 3 |
| Entry | 1 | 100 | | | 1 | 99 | 1 | | 1 | 100 | | | 1 | 100 | | |
| | 2 | 4 | 96 | | 2 | 4 | 79 | 17 | 2 | 17 | 70 | 13 | 2 | 21 | 79 | |
| | 3 | | | 100 | 3 | | 6 | 94 | 3 | | | 100 | 3 | | | 100 |
| | | E | | | | S | | | | W | | | | N | | |
| | C | 1 | 2 | 3 | C | 1 | 2 | 3 | C | 1 | 2 | 3 | C | 1 | 2 | 3 |
| Exit | 1 | 100 | | | 1 | 100 | | | 1 | 100 | | | 1 | 100 | | |
| | 2 | 12 | 88 | | 2 | 33 | 63 | 4 | 2 | 31 | 69 | | 2 | 9 | 91 | |
| | 3 | 10 | | 90 | 3 | | | 100 | 3 | | | 100 | 3 | | 4 | 96 |
| | | $E_l$ | | | | $E_r$ | | | | $E_s$ | | | | $S_l$ | | |
| | C | 1 | 2 | 3 | C | 1 | 2 | 3 | C | 1 | 2 | 3 | C | 1 | 2 | 3 |
| | 1 | 100 | | | 1 | 100 | | | 1 | 100 | | | 1 | 100 | | |
| | 2 | 4 | 96 | | 2 | 41 | 59 | | 2 | 7 | 93 | | 2 | 20 | 80 | |
| | 3 | | | 100 | 3 | | | 100 | 3 | | | 100 | 3 | | | 100 |
| EDD | | $S_r$ | | | | $S_s$ | | | | $W_l$ | | | | $W_s$ | | |
| | C | 1 | 2 | 3 | C | 1 | 2 | 3 | C | 1 | 2 | 3 | C | 1 | 2 | 3 |
| | 1 | 100 | | | 1 | 99 | 1 | | 1 | 100 | | | 1 | 100 | | |
| | 2 | | 100 | | 2 | 9 | 91 | | 2 | 3 | 97 | | 2 | 10 | 90 | |
| | 3 | | | 100 | 3 | | | 100 | 3 | | | 100 | 3 | | | 100 |
| | | $N_l$ | | | | $N_r$ | | | | $N_s$ | | | | *Average* | | |
| | C | 1 | 2 | 3 | C | 1 | 2 | 3 | C | 1 | 2 | 3 | C | 1 | 2 | 3 |

| | | 1 | 100 | | | 1 | 100 | | | 1 | 100 | | | 1 | 100 | |
|---|---|---|---|---|---|---|---|---|---|---|---|---|---|---|---|---|
| | | 2 | 17 | 83 | | 2 | 27 | 73 | | 2 | 22 | 73 | | 2 | 15 | 83 | 2 |
| | | 3 | | 100 | 3 | | | 100 | 3 | | | 100 | 3 | 1 | | 99 |

Tables 7 and 8 list the classification accuracies by using the proposed method with RBF kernel for the AM and PM sessions, respectively. From the results, we can see that the proposed method with linear kernels as shown in Tables 3 and 4 has better classification accuracies than that with the RBF kernel (average 94.26% (AM) and 89.82% (PM) classification accuracies). It means that the liner kernel is superior to the RBF kernel in the SVM model for the application of AIC with discriminative features.

Table 7. Classification results based on the proposed method in the AM sessions (%)

| | Direction | 1 | 2 | 3 | 4 | 5 | Average |
|---|---|---|---|---|---|---|---|
| **Entry** | E | 94.20 | 95.65 | 94.20 | 92.75 | 91.30 | 93.62 |
| | S | 97.10 | 91.30 | 94.20 | 92.75 | 94.20 | 93.9 |
| | W | 98.55 | 98.55 | 95.65 | 94.20 | 98.55 | 97.10 |
| | N | 92.75 | 94.20 | 94.20 | 92.75 | 89.86 | 92.75 |
| **Exit** | E | 91.30 | 97.10 | 98.55 | 95.65 | 94.20 | 95.36 |
| | S | 92.75 | 91.30 | 91.30 | 92.75 | 97.10 | 93.04 |
| | W | 95.65 | 94.20 | 97.10 | 95.65 | 95.65 | 95.65 |
| | N | 94.20 | 91.30 | 92.75 | 94.20 | 97.10 | 93.91 |
| **EDD** | $E_l$ | 97.10 | 95.65 | 95.65 | 91.30 | 98.55 | 95.65 |
| | $E_r$ | 89.86 | 94.20 | 92.75 | 91.30 | 88.41 | 91.30 |
| | $E_s$ | 97.10 | 92.75 | 94.20 | 98.55 | 89.86 | 94.49 |
| | $S_l$ | 95.65 | 98.55 | 97.10 | 95.65 | 94.20 | 96.23 |
| | $S_r$ | 97.10 | 94.20 | 92.75 | 92.75 | 91.30 | 93.62 |
| | $S_s$ | 92.75 | 91.30 | 97.10 | 94.20 | 94.20 | 93.91 |
| | $W_l$ | 94.20 | 97.10 | 97.10 | 86.96 | 94.20 | 93.91 |
| | $W_s$ | 95.65 | 94.20 | 89.86 | 92.75 | 94.20 | 93.33 |
| | $N_l$ | 94.20 | 94.20 | 89.86 | 94.20 | 95.65 | 93.62 |
| | $N_r$ | 94.20 | 95.65 | 97.10 | 94.20 | 91.30 | 94.49 |
| | $N_s$ | 95.65 | 94.20 | 92.75 | 95.65 | 97.10 | 95.07 |
| | | | | Average | | | **94.26** |

Table 8. Classification results based on the proposed method in the PM sessions (%)

|  | Direction | 1 | 2 | 3 | 4 | 5 | Average |
|---|---|---|---|---|---|---|---|
| **Entry** | E | 88.40 | 89.86 | 91.30 | 89.86 | 91.30 | 90.14 |
|  | S | 88.41 | 88.41 | 97.10 | 94.20 | 91.30 | 91.88 |
|  | W | 89.86 | 85.51 | 91.30 | 91.30 | 86.96 | 88.99 |
|  | N | 88.41 | 94.20 | 81.16 | 81.16 | 86.96 | 86.38 |
| **Exit** | E | 88.41 | 86.96 | 91.30 | 91.30 | 89.86 | 89.57 |
|  | S | 89.86 | 91.30 | 86.96 | 92.75 | 86.96 | 89.57 |
|  | W | 92.75 | 88.41 | 89.86 | 91.30 | 92.75 | 91.01 |
|  | N | 88.41 | 94.20 | 89.86 | 84.06 | 86.96 | 88.70 |
| **EDD** | $E_l$ | 89.86 | 88.41 | 92.75 | 88.41 | 91.30 | 90.14 |
|  | $E_r$ | 84.06 | 92.75 | 88.41 | 92.75 | 89.86 | 89.57 |
|  | $E_s$ | 89.86 | 85.51 | 88.41 | 88.41 | 91.30 | 88.70 |
|  | $S_l$ | 95.65 | 92.75 | 95.65 | 89.86 | 92.75 | 93.33 |
|  | $S_r$ | 92.75 | 82.61 | 89.86 | 94.20 | 94.20 | 90.72 |
|  | $S_s$ | 88.41 | 89.86 | 92.75 | 91.30 | 85.51 | 89.57 |
|  | $W_l$ | 94.20 | 84.06 | 95.65 | 94.20 | 88.41 | 91.30 |
|  | $W_s$ | 89.86 | 91.30 | 92.75 | 91.30 | 89.86 | 91.01 |
|  | $N_l$ | 89.86 | 89.86 | 86.96 | 88.41 | 86.96 | 88.41 |
|  | $N_r$ | 91.30 | 88.41 | 89.86 | 85.51 | 86.96 | 88.41 |
|  | $N_s$ | 88.41 | 86.96 | 88.41 | 89.86 | 92.75 | 89.28 |
|  | Average |  |  |  |  |  | 89.82 |

4.2 Relationship between classification accuracy and PCA dimension

In order to have a discriminative feature to represent a ST signal, PCA is used to reduce the dimension of feature vector of a traffic signal before training. To study the variation of classification accuracy with the dimension of feature vectors, 17 groups of experiments with different dimensions of PCA feature vectors were performed. To produce robust results, a 5-fold cross-validation approach was used in each group of experiment. Fig. 8 shows the variation of averaging classification accuracy of the 19 traffic directions with respect to different dimensions of PCA feature vectors. From Fig. 8, an optimal classification performance is achieved when the dimension of PCA feature vector is set to 25 and 40 for the AM and PM sessions, respectively.

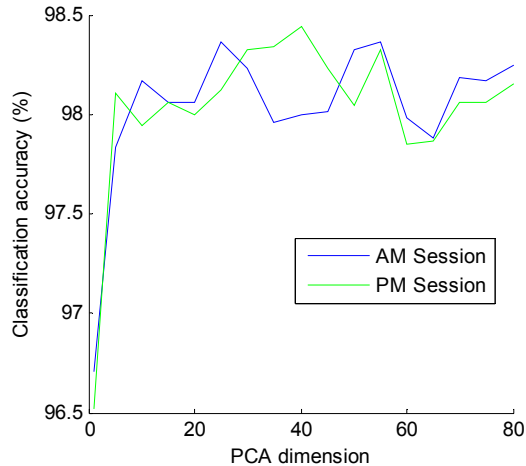

Fig.8 Variation of classification accuracy with different dimensions of PCA feature vector

5. Conclusion

In this paper, an AIC method has been presented to identify traffic states in big traffic data. In the proposed method, traffic data are firstly represented as ST signals, and a set of simulation data are thereby generated to construct an extended big traffic database. A Mean-Shift filter is then used to extract the features of ST signals. In order to decrease training complexity, the PCA is adopted to reduce the feature dimension and extract discriminative representation of signal features. In the training stage, a hybrid model combines an adaptive boosting classifier and a SVM learning to detect outliers and further to identify the categories of outliers. Experimental results show that the proposed method achieves high classification accuracy.

In the future, we would explore the differences of traffic characteristics in different traffic directions. Based on the analysis, more discriminative feataures should be extracted for training and testing stages. It is expected that the slight traffic jams would also be discerned with the higher accuracy. Another research direction is to estimate the optical flow of objects of interest from camera videos instead of the ST signals. AIC would be performed based on the distribution of optical flow fields.

## Acknowledgment

This research is supported by the grants of Hong Kong RGC GRF: 12201814 and HKBU FRG/14-15/054.

[22] X. Jin, R. L. Cheu, and D. Srinivasan, "Development and adaptation of constructive probabilistic neural network in freeway incident detection," Transportation Research Part C: Emerging Technologies, vol. 10, pp. 121-147, 2002.

[23] R. L. Cheu, D. Srinivasan, and E. T. Teh, "Support vector machine models for freeway incident detection," in Intelligent Transportation Systems, 2003. Proceedings. 2003 IEEE, 2003, pp. 238-243.

[24] F. Yuan and R. L. Cheu, "Incident detection using support vector machines," Transportation Research Part C: Emerging Technologies, vol. 11, pp. 309-328, 2003.

[25] Z. Zhou and L.-y. Zhou, "An Automatic Incident of Freeway Detection Algorithm Based on Support Vector Machine," in Intelligence Information Processing and Trusted Computing (IPTC), 2010 International Symposium on, 2010, pp. 543-546.

[26] M. Kearns, "Thoughts on hypothesis boosting," Unpublished manuscript, vol. 45, p. 105, 1988.

[27] M. Kearns and L. Valiant, "Cryptographic limitations on learning Boolean formulae and finite automata," Journal of the ACM (JACM), vol. 41, pp. 67-95, 1994.

[28] R. E. Schapire, "The strength of weak learnability," Machine learning, vol. 5, pp. 197-227, 1990.

[29] B. Wu, H. Ai, C. Huang, and S. Lao, "Fast rotation invariant multi-view face detection based on real adaboost," in Automatic Face and Gesture Recognition, 2004. Proceedings. Sixth IEEE International Conference on, 2004, pp. 79-84.

[30] R. Lienhart, A. Kuranov, and V. Pisarevsky, "Empirical analysis of detection cascades of boosted classifiers for rapid object detection," in Pattern Recognition, ed: Springer, 2003, pp. 297-304.

[31] V. N. Vapnik and V. Vapnik, Statistical learning theory vol. 1: Wiley New York, 1998.

[32] F. Hu, X. Liu, J. Dai, and H. Yu, "A Novel Algorithm for Imbalance Data Classification Based on Neighborhood Hypergraph," The Scientific World Journal, vol. 2014, 2014.

[33] J. Qin and N. H. Yung, "Feature fusion within local region using localized maximum-margin learning for scene categorization," Pattern Recognition, vol. 45, pp. 1671-1683, 2012.

[34] M. M. Adankon and M. Cheriet, "Model selection for the LS-SVM. Application to handwriting recognition," Pattern Recognition, vol. 42, pp. 3264-3270, 2009.